%% file: egpaper_final.tex
\documentclass[10pt,twocolumn,letterpaper]{article}

\usepackage{3dv}
\usepackage{times}
\usepackage{epsfig}
\usepackage{graphicx}
\usepackage{amsmath}
\usepackage{amssymb}
\usepackage{pifont}
\usepackage{gensymb}
\usepackage{bbm}
\usepackage{booktabs}
\usepackage[ruled,vlined]{algorithm2e}
\usepackage{xcolor}
\usepackage{subcaption}
\usepackage{enumitem}
\usepackage{multirow}

\usepackage[pagebackref=true,breaklinks=true,letterpaper=true,colorlinks,bookmarks=false]{hyperref}

\threedvfinalcopy 

\def\threedvPaperID{078} 

\ifthreedvfinal\fi
\begin{document}

\title{SF-UDA$^{3D}$: Source-Free Unsupervised Domain Adaptation \\ for LiDAR-Based 3D Object Detection}

\author{Cristiano Saltori\\
University of Trento\\
{\tt\small cristiano.saltori@unitn.it}
\and
St\'{e}phane Lathuili\'{e}re\\
LTCI, T\'{e}l\'{e}com Paris, Institut Polytechnique de Paris\\
{\tt\small stephane.lathuiliere@telecom-paris.fr}
\and
Nicu Sebe\\
University of Trento\\
Huawei Research\\
{\tt\small niculae.sebe@unitn.it}
\and
Elisa Ricci\\
University of Trento\\
Fondazione Bruno Kessler\\
{\tt\small eliricci@fbk.eu}

\and
Fabio Galasso\\
Sapienza University of Rome\\
{\tt\small galasso@di.uniroma1.it}
}

\maketitle

\begin{abstract}

3D object detectors based only on LiDAR point clouds hold the state-of-the-art on modern street-view benchmarks.
However, LiDAR-based detectors poorly generalize across domains due to domain shift. In the case of LiDAR, in fact, domain shift is not only due to changes in the environment and in the object appearances, as for visual data from RGB cameras, but is also related to the geometry of the point clouds (e.g., point density variations).
This paper proposes SF-UDA$^{3D}$, the first Source-Free Unsupervised Domain Adaptation (SF-UDA) framework to domain-adapt the state-of-the-art PointRCNN 3D detector to target domains for which we have no annotations (unsupervised), neither we hold images nor annotations of the source domain (source-free). SF-UDA$^{3D}$ is novel on both aspects. Our approach is based on pseudo-annotations, reversible scale-transformations and motion coherency.
SF-UDA$^{3D}$ outperforms both previous domain adaptation techniques based on features alignment and state-of-the-art 3D object detection methods which additionally use few-shot target annotations or target annotation statistics. This is demonstrated by extensive experiments on two large-scale datasets, i.e., KITTI and nuScenes. 

\end{abstract}

\input{introduction}

\input{related_works}

\input{method}

\input{experiments}
\input{ablation}
\input{conclusions}
\vspace{-0.3cm}
\section*{Acknowledgment}

\noindent The work was partially supported by OSRAM GmbH and was carried out in the Vision and Learning joint laboratory of FBK and UNITN. We thank the CARITRO Deep Learning laboratory of ProM Facility for the granted GPU time.

{\small
\bibliographystyle{ieee}
\bibliography{egbib}
}

\end{document}

%% file: introduction.tex
\vspace{-0.3cm}
\section{Introduction}

LiDAR is one of the key sensors for the longer-term autonomy of cars~\cite{autoCarSur19}. It is a native 3D sensor, which reads up to hundreds of meters, providing point clouds. Further to the proliferation of LiDAR companies, the efficacy of LiDAR is proven by the fact that LiDAR-only-based detectors are robust and accurate~\cite{yan2018second, shi2019pointrcnn, xu2018pointfusion, lang2019pointpillars} and provide state-of-the-art performance, currently held by PointRCNN~\cite{shi2019pointrcnn}.

\begin{figure}[t]
    \centering
    \includegraphics[width=\columnwidth]{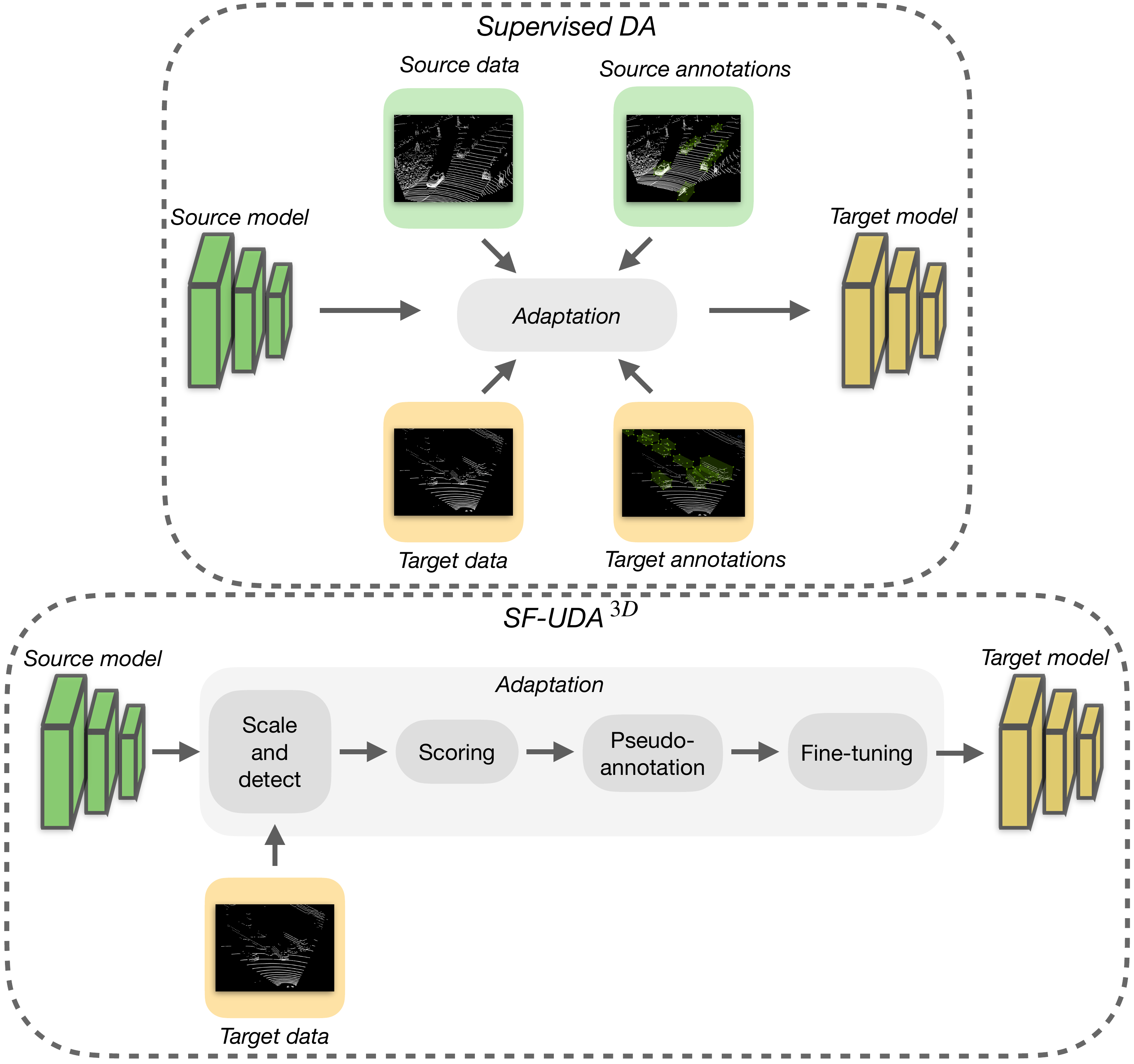}
    \caption{Existing supervised DA methods for LiDAR-based 3D detection \cite{wang2020train} require both source and target data and annotations to adapt a pre-trained deep model to a target domain. Differently, leveraging on pseudo-annotations, reversible scale-transformations and motion coherency, SF-UDA$^{3D}$ adapts a pre-trained source network by using only unlabeled target data.}
    \label{fig:teaser}
    \vspace{-0.7cm}
\end{figure}

LiDAR-based detectors, however, are prone to domain shift issues that may be more serious than for their RGB counterparts \cite{wang2020train}. As in the RGB case, domain shift may be due to environmental changes (\textit{e.g.} data collected in different cities and weather conditions) or to appearances variations of specific objects (\textit{e.g.} car shapes and sizes may vary among different countries). Additionally, performance of LiDAR-based models significantly depends on the density of the LiDAR point cloud, spatial resolution and ranges. 

To address this problem, in this paper we propose SF-UDA$^{3D}$, the first Source-Free Unsupervised Domain Adaptation framework for LiDAR-based 3D detection. The proposed technique features the case where the 3D detector is applied \textit{e.g.}\ in a different country, with differences in the local cars and roads, as well as to imagery acquired with different LiDAR sensors. Our approach is an \textit{unsupervised} DA method because it does not require any annotation in the target domain and it is \textit{source-free} because we assume that only the 3D detector trained on source data is available, while we do not have access to the source annotations and data (see Fig. \ref{fig:teaser}). Both aspects are novel. To the best of our knowledge, in fact, despite its practical relevance, the problem of building LiDAR-based 3D detectors which are robust to domain shift has only recently been addressed in \cite{wang2020train}. However, the method proposed in \cite{wang2020train} assumes the availability of both source and target domain data and annotations. In this paper, we argue that this assumption is rarely satisfied in many real-world applications, where we may have only access to a pre-trained detector and it may be hard or even impossible to acquire target data annotations.

SF-UDA$^{3D}$ considers the PointRCNN~\cite{shi2019pointrcnn} architecture and is based on the pseudo-annotation of the target unlabelled dataset by means of reversible scale-transformations and motion coherency. In details, SF-UDA$^{3D}$ annotates the unlabelled target data at multiple scales by a PointRCNN model pre-trained on the unavailable source dataset. Then, it assesses the quality of the annotations by scoring the resulting detection tracks by means of an unsupervised coherency metric (Mean Volume Variation). Subsequently, it reverses the scale-transformations, aggregates the best detection labels by confidence and finally fine-tunes Point RCNN. 
In spite of its simplicity, SF-UDA$^{3D}$ surpasses in performance state-of-the-art domain adaptation methods for 3D object detection which require target data annotations or annotation statistics~\cite{wang2020train}. Our algorithm also 
outperforms previous source-free general purpose unsupervised domain adaptation methods \cite{li2016revisiting} by 30\%. Overall, we fill in 66\% of the gap between the source and target 3D detector when adapting from nuScenes to KITTI, and 30\% in the much harder case of KITTI to nuScenes.

To summarize, our main contributions are as follows:
\setlist{nolistsep}
\begin{itemize}[noitemsep]
    \item  We propose to study a novel problem, \textit{i.e.} how to build LiDAR-based 3D detectors robust to domain shift when (i) we do not have access to source data and annotations and only a source pre-trained model is available, and (ii) no annotations are provided in the target domain, \textit{i.e.} we are in an unsupervised domain adaptation setting; 
    \item We propose SF-UDA$^{3D}$, a novel approach for source-free unsupervised domain adaptation which empowers the state-of-the-art PointRCNN~\cite{shi2019pointrcnn} architecture by means of pseudo-annotations, reversible scale-transformations and motion coherency;
    \item We evaluate the proposed SF-UDA$^{3D}$ against relevant approaches and we show that our method outperforms both previous source-free feature-based domain adaptation methods~\cite{li2016revisiting} and, notably, state-of-the-art adaptation approaches for LiDAR-based 3D detection, although they additionally use few-shot target annotations or target annotation statistics~\cite{wang2020train}.
\end{itemize}



%% file: related_works.tex
\vspace{-0.3cm}
\section{Related Works}

This work relates to LiDAR-based 3D detection, unsupervised 2D domain adaptation and 3D domain adaptation, which is currently at least weakly-supervised.

\noindent\textbf{LiDAR-based 3D Object Detection.}
First successful 3D object detectors were two-stage detectors and employed RGB images to propose object bounding-boxes (with a region proposal network RPN), prior to a 3D classification branch (termed RCNN)~\cite{chen2017multi, qi2018frustum, wang2019frustum, xu2018pointfusion, ku2018joint}.
More recently, better performance has been obtained by object proposal stages also based on 3D point clouds. This is the case of VoxelNet~\cite{zhou2018voxelnet}, SECOND~\cite{yan2018second} and PointPillar~\cite{lang2019pointpillars} which encoded the 3D point clouds into voxel or pillars and used by the RPN~\cite{zhou2018voxelnet, yan2018second} or by a 2D detections head~\cite{lang2019pointpillars} for generating object proposals. The state-of-the-art LiDAR-based 3D detector is PointRCNN~\cite{shi2019pointrcnn}, which avoids pseudo-images and the consequent quantization. It processes instead the point clouds in 3D, employing PointNet++~\cite{qi2017pointnet++} 3D features both in the RPN and RCNN stages. Here we adopt PointRCNN both for its performance and robustness. 

\noindent\textbf{Unsupervised Domain Adaptation (UDA) for 2D Object Detection.}
There is a large body of literature on 2D UDA detection~\cite{tzeng2018splat, kim2019diversify, wang2019towards, wang2019few, he2019multi, saleh2019domain, shen2019scl, cao2019pedestrian}, which may be divided into three types of approaches~\cite{li2020deep}. The first tackles the domain shift by enforcing domain confusion with adversarial training \cite{chen2018domain, zhu2019adapting, saito2019strong, zhuang2020ifan}, either with image-domain classifiers only~\cite{chen2018domain} or by additional region-level alignments~\cite{zhu2019adapting}, possibly leveraging multiple scales and semantic information~\cite{zhuang2020ifan}.
The second type of approaches exploits image-to-image translation between the source and target-domain data~\cite{arruda2019cross, lin2019cross, liu2018ir2vi}. Some methods also add adversarial losses, which makes them hybrid
between the two types~\cite{shan2019pixel, kim2019diversify, kim2019self, rodriguez2019domain, hsu2020progressive}.
A third type of methods, more closely related to our work, has considered self-training on target pseudo-labels~\cite{khodabandeh2019robust, cai2019exploring, roychowdhury2019automatic}. In \cite{khodabandeh2019robust}, noisy pseudo-annotations are obtained with the source model and refined by a joint source-target classifier; then, the target-adapted model is trained from scratch on the pseudo-labels. Differently, \cite{cai2019exploring} adopts a teacher-student framework and similarly leverages the pseudo-labelling by a mean teacher, \textit{i.e.} using the source supervised loss. In fact these techniques assume that the source dataset and labels are available during the domain adaptation phase, while our SF-UDA$^{3D}$ operates in a source-free setting.
More recently, in the context of 2D detection, \cite{roychowdhury2019automatic} has adapted a source-trained model in a source-free manner by temporal consistency and knowledge distillation~\cite{hinton2015distilling}. First, they labeled missed detections by tracking in the target dataset, then they distilled the soft-labels and re-train the detector.
In this work, we also use temporal consistency, but we employ tracking to determine the scales which best transform the target point clouds for 3D pseudo-labelling.

\noindent\textbf{Domain Adaptation (DA) from 3D LiDAR data.}
To the best of our knowledge, unsupervised domain adaptation from LiDAR data has been so far only researched for classification~\cite{qin2019pointdan,achituve2020self} and segmentation tasks~\cite{wu2019squeezesegv2,zhao2019multi}. In all cases, the algorithms assume that the source images and labels are available. This differs from SF-UDA$^{3D}$, which is instead a source-free pipeline.
There are currently two works of DA for 3D detection in LiDAR data. The first~\cite{wang2019range} aligns the global and local features by adversarial training, therefore \cite{wang2019range} requiring supervision, \textit{i.e.}\ annotation of the target data, and it is not source-free. The second work~\cite{wang2020train} evaluates both a few-shot DA, where a few annotations are given for the target dataset, and a weakly-supervised DA, where aggregated target statistics about the car scales are provided for adaptation via fine-tuning on the scale-transformed source annotations. Our method also leverages on scale-transformations but of the entire input point clouds and it is source-free and completely unsupervised, as it estimates the target scales via temporal coherency. Additionally, it aggregates pseudo-annotations from several estimated relevant scales, which makes a substantial difference in performance against the single output transformation or few-shot fine-tuning in~\cite{wang2020train}, as we quantitatively evaluate in Sec.~\ref{sec:experiments}.

%% file: method.tex
\begin{figure*}[ht]
    \centering
    \includegraphics[scale=0.65]{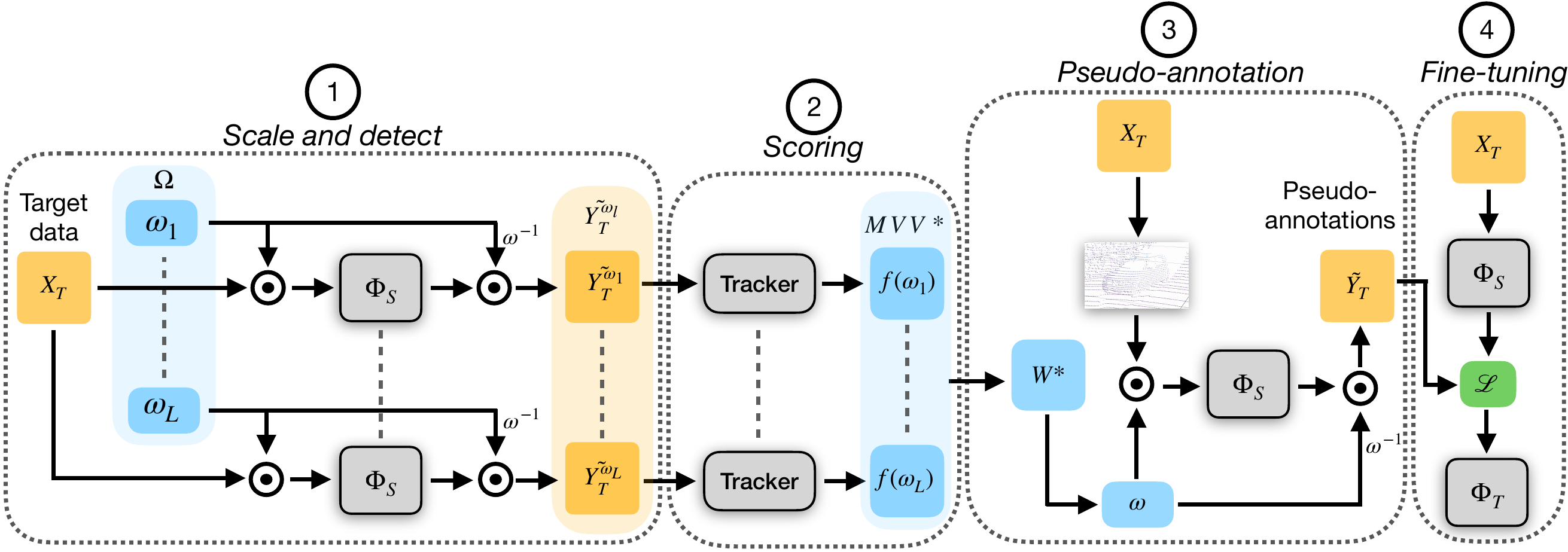}
    \caption{\textbf{Full SF-UDA$^{3D}$ pipeline overview.} Given a scaling solution space $\Omega$, in the first step detections over target sequences are obtained by scaling input data by $\omega$ and by re-scaling predictions by $1 / \omega$. Next, time consistency of each sequence is used through a tracker to score each solution. During the third stage, scores are used to identify the best scaling interval $W^*$ and pseudo-annotations are obtained over multiple iterations with the same procedure of step one and are merged through NMS. Finally, we obtain the target adapted model $\Phi_\mathcal{T}$ by fine-tuning the source model over target data and pseudo-annotations.}
    \label{fig:full}
\end{figure*}

\vspace{-0.2cm}
\section{SF-UDA$^{3D}$}
\label{sec:method}

In this section, we first present the problem statement and an overview of the proposed approach for domain adaptation, then we describe it in details in the subsections.
\vspace{-0.1cm}
\subsection{Problem Statement}
Given a 3D detection model $\Phi_S$ trained on a source dataset $\mathcal{X}_\mathcal{S} = \{x^n_S\}_{1<n<N}$ with source annotations $\mathcal{Y}_{\mathcal{S}} = \{y^n_S\}_{1<n<N}$, the goal of UDA in 3D detection is to obtain a target adapted 3D detector $\Phi_T$ by exploiting a target dataset $\mathcal{X}_\mathcal{T} = \{x^m_T\}_{1<m<M}$ without ground-truth annotations  $\mathcal{Y}_\mathcal{T} = \{y^m_T\}_{1<m<M}$. In this work, we consider the challenging scenario of Source-Free Unsupervised Domain Adaptation (SF-UDA), where the source data $\mathcal{X}_{\mathcal{S}}$, the source annotations $\mathcal{Y}_{\mathcal{S}}$ and the target annotations $\mathcal{Y}_\mathcal{T}$ are not available at adaption time, namely, when training the target model $\Phi_T$.

When the LiDAR sensors differ across source and target domains, the geometries of the point clouds are different.
Assume that the source and target point clouds are sampled by respective generating probability distributions $P(\mathcal{X})$, \textit{i.e.} $P(\mathcal{X}_{\mathcal{S}})\neq P(\mathcal{X}_{\mathcal{T}})$. We illustrate in Sec.~\ref{sec:experiments} that this is especially the case for point could densities in the nuScenes~\cite{nuscenes2019} and KITTI~\cite{geiger2012we} datasets considered in this work. When the source and target datasets are acquired across different domains, \textit{e.g.} countries, the ground-truth annotations also differ, \textit{e.g.} since the shapes of cars are also different. Assuming that annotations are sampled from generating probability distributions, these would differ, \textit{i.e.} $P(\mathcal{Y}_{\mathcal{S}})\neq P(\mathcal{Y}_{\mathcal{T}})$. This is the case of the nuScenes~\cite{nuscenes2019} and KITTI~\cite{geiger2012we} datasets, acquired respectively in USA/Singapore and Germany.
The same discrepancy should be reflected in the 3D detector output spaces, trained to mimic the ground-truth annotations.

In this work, we propose to align the annotation-scale distributions $P(\mathcal{Y}_{\mathcal{S}})$ and $P(\mathcal{Y}_{\mathcal{T}})$ by scale-transformation parameters, which we estimate by temporal coherency. Furthermore, we account for the misalignment of the point cloud distributions $P(\mathcal{X}_{\mathcal{S}})$ and $P(\mathcal{X}_{\mathcal{T}})$ by fine-tuning the source model $\Phi_S$ on the pseudo-annotated target point cloud.

\vspace{-0.2cm}
\subsection{Proposed Method}

We propose a four-stage pipeline for adaptation, detailed in Secs.~\ref{sec:scaling}-\ref{sec:annotation} and illustrated in Fig.~\ref{fig:full}.
Firstly, we detect objects with the source model $\Phi_S$ over a set of scaled target point clouds (Sec.~\ref{sec:scaling}). Secondly, we score the detections by a tracker for temporal consistency (Sec.~\ref{sec:tracking}). Thirdly, we aggregate detections at the best scales and lastly we fine-tune the source-model on the pseudo-annotated target point-cloud (Sec.~\ref{sec:annotation}). 

\vspace{-0.4cm}
\subsubsection{Scale and detect}
\label{sec:scaling}

In the \emph{Scale-and-detect} stage, we consider a set of $L$ scaling parameters $\Omega=[\omega_1, ..., \omega_{L}]$ where each $\omega_l=(\omega_x, \omega_y, \omega_z)\in{\mathbf{R}^+}^3$ parametrizes the scaling transformation along the 3D axes. To specify $\Omega$, we use a regular grid over the intervals centered around 1: $[1-\epsilon, 1+\epsilon]^3$ with $\epsilon>0$. 
For each $\omega_l$, we generate a transformed version of the dataset $\mathcal{X}_\mathcal{T}^{\omega_l}$ by re-scaling each sample in $\mathcal{X}_\mathcal{T}$. Then, we employ the source object detector $\Phi_S$ on every sample of $\mathcal{X}_\mathcal{T}^{\omega_l}$ obtaining detections $\tilde{\mathcal{Y}}_\mathcal{T}^{\omega_l}$. Finally, to have detections $\tilde{\mathcal{Y}}_\mathcal{T}^{\omega_l}$ in the original target 3D space, we re-scale the detections by multiplying the position and dimension values by $(1/\omega_x, 1/\omega_y, 1/\omega_z)$.  Besides, we note that this re-scaling step is required to obtain detections in the same 3D space and to allow a fair temporal consistency comparison. In all our experiments, we employ the PointRCNN detector \cite{shi2019pointrcnn} since it recently obtained the state-of-the-art performance in object detection benchmarks.

\vspace{-0.4cm}
\subsubsection{Scale scoring with Temporal consistency}
\label{sec:tracking}
\begin{figure}[h]
    \centering
    \includegraphics[width=\columnwidth]{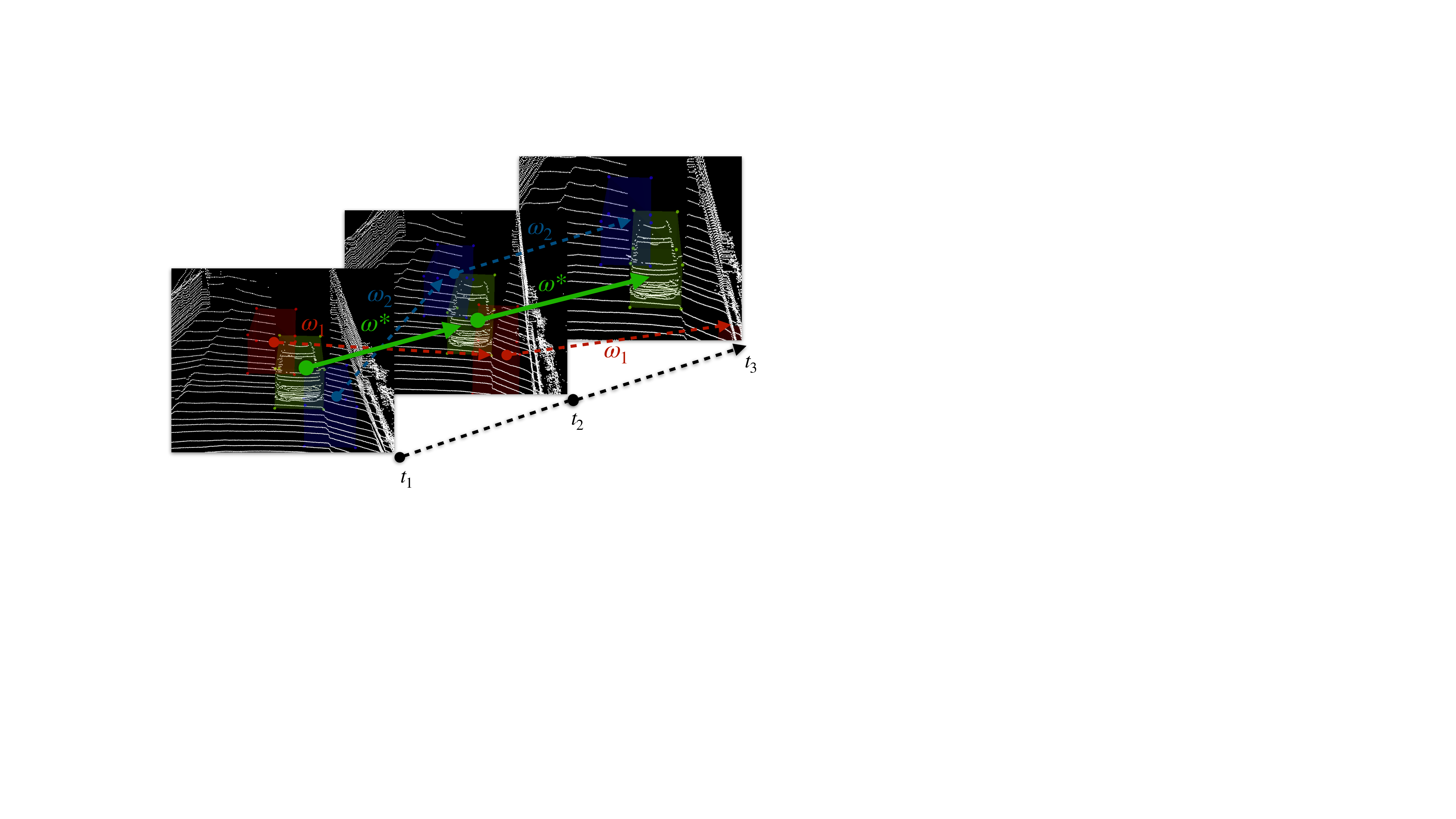}
    \caption{Given multiple possible scales $\omega$, SF-UDA$^{3D}$ selects the best  $\omega^*$ as the one generating the most time consistent detections.}
    \label{fig:scale_tracks}
\end{figure}

To identify the quality of the estimated detections $\tilde{\mathcal{Y}}_\mathcal{T}^{\omega_l}$ , we leverage the temporal consistency of detections between sequential frames (see Fig. \ref{fig:scale_tracks}). We propose to use a tracker and to evaluate the stability of its prediction to \textit{score the detection quality}.
More specifically, we run a state-of-the-art tracking-by-detection pipeline \cite{weng2019baseline}. For a given sequence $V$, we assume to obtain $J$ tracks. Considering the tracked object with the index $j<J$, its track can be defined as lists of $T_j$ consecutive bounding-boxes $B_j=[b_{t_j}, ..., b_{t_j+T_j}]$, where $t_j$ denotes the frame index where the object appears, and each $b_t$ is the 3D bounding-box dimensions and locations at time $t$.
Inspired by \cite{xu2016libsvx}, we employ the Mean Volume Variation (MVV) between consecutive detections as scoring function:
\vspace{-0.3cm}
\begin{equation}
    MVV(V) = \frac{1}{J} \sum_{j=1}^J\sqrt{\frac{\sum_{t=t_j}^{t_j+T_j} (v_j^t - \Bar{v_j})^2}{T_j-1}}
    \label{eq:mvv}
\end{equation}

where $v_j^t$ is the bounding box volume of the $j$-th track at time $t$ and $\Bar{v_j}$ is the mean volume of the bounding boxes in $B_j$. The intuition behind this scoring function is that, assuming an optimal detector and rigid objects, the bounding-box volume would be constant. Therefore, a good detector would predict detections with a stable volume leading to a small \emph{MVV} value. 
Importantly, we observed that tracks that last less than 5 frames may be false-positives, and therefore, we propose to treat these tracks differently. More precisely, we introduce a penalty term $H^*$ in the \emph{MVV} score for every sequence without tracks longer than 5 frames. 
Our robust version of the  MVV score can be written as:
\vspace{-0.2cm}
\begin{equation}
    MVV^*(V) = \begin{cases} MVV(V), & \mbox{$J\neq0$} \\ H^*, & \mbox{if no tracks} \end{cases}
    \label{eq:mvv_penalty}
\end{equation}

Finally, as previously mentioned, we are interested in scoring each scaling solution over all the target training sequences, therefore we consider as the scoring function $f(\omega_i)$ for $\omega_i$ the mean of $MVV^*$ over all the target sequences.
\vspace{-0.4cm}
\vspace{-0.8cm}

    
        


    
    
\vspace{-0.1cm}
\subsubsection{Pseudo-annotation and Fine-tuning}
\label{sec:annotation}

Once each scaling parameters in $\Omega$ is scored, we proceed with \emph{Pseudo-annotation}. In a first \emph{single-scale} (\emph{SS}) approach, we only consider the scaling parameter corresponding to the lowest MVV$^*$, \textit{i.e.}\ the best scale, referred to as $\omega^*$ (see Fig. \ref{fig:scale_tracks}).
In this approach, the detections $\tilde{\mathcal{Y}}_\mathcal{T}^{\omega*}$ at scale $\omega^*$ from the first stage of our pipeline are considered as pseudo-annotations. We confirm in Sec.~\ref{sec:experiments} that the identified best scale provides the best single-scale results.
We also propose a \textit{multi-scale} (\emph{MS}) approach, which combines the top-$K$ best scaling parameters to improve the pseudo-annotations. We name $\Omega^*$ this set of best scales and we use it to determine a best scaling interval $W^*=W_x^*\times W_y^*\times W_z^*$, as follows:
\vspace{-0.3cm}
\begin{equation}
      \forall a \in [x,y,z], W_a^* = [\min_{\omega \in \Omega^*}(\omega_a), \max_{\omega \in \Omega^*}(\omega_a)]
\label{eq:interval}
\end{equation}
Given $W^*$, for every frame of the target dataset, we sample a random $\omega$ in $W^*$. The point cloud is scaled using $\omega$. Then, we use the source model $\Phi_S$ to obtain 3D detections. Finally, we re-scale the predictions with $1/\omega$ similarly to the \emph{scale-and-detect} stage, to obtain bounding boxes in the original point-cloud space.

Sampling scales is repeated several times, the resulting 3D detections are collected and aggregated by non-maximal-suppression (NMS), yielding the final pseudo-labels $\tilde{\mathcal{Y}}_\mathcal{T}$. Sampling the scales multiple times is beneficial, possibly for two reasons: \text{(i)}\ cars within a point cloud occur at multiple scales and multiple sampling may find more of them; \text{(ii)}\ PointRCNN randomly sub-samples the input point cloud to obtain a constant number of input points and having multiple scale samples ensures more robustness against this randomness. 
Note that PointRCNN provides detections with confidence scores, which we find beneficial to threshold increasingly from low to high values. 
In other words, at early steps we use a low threshold to increase recall and include also low confidence detections. Later, we raise the threshold and only consider detections with a higher confidence.
Finally, we merge pseudo-annotations from different steps through NMS and fine-tune the source model $\Phi_S$ by using $\tilde{\mathcal{Y}}_\mathcal{T}$ as annotations.
\vspace{-0.2cm}

%% file: experiments.tex
\section{Experiments}
\label{sec:experiments}

In this section, we present the experimental evaluation of SF-UDA$^{3D}$ on two modern large-scale benchmarks of KITTI~\cite{geiger2012we, geiger2013vision} and nuScenes~\cite{nuscenes2019} against state-of-the-art methods, albeit none of them is source-free and unsupervised. Then, we conduct a thorough ablation study of the proposed framework. In the following, we introduce benchmarks and metrics.


\input{datasets}

\noindent\textbf{Metrics.}
For KITTI, we adopt the official metrics of \cite{geiger2012we, geiger2013vision} and the classification into easy/moderate/hard detections, according to the visibility of the objects. In more details, we report the Average Precision (AP) over the 3D Intersection over the union (IoU) with a IoU threshold of $0.7$. The final average (Avg) AP is obtained by averaging over the three difficulty categories.\\
For nuScenes we consider the official metrics~\cite{nuscenes2019}. In our experiments we consider the center-based definition of AP as used in the official nuScenes benchmark \cite{nuscenes2019} and report both the AP at each of the four different distance thresholds of $[0.5, 1.0, 2.0, 4.0]$ meters and the final average (Avg) over the four thresholds.



\begin{table}[t]
    \centering
    \resizebox{0.9\columnwidth}{!}{
    \begin{tabular}{c|cccccccc}
      \toprule
        \textbf{Method} & \textbf{Easy} & \textbf{Moderate} & \textbf{Hard} & \textbf{Avg-AP}\\ \midrule 
        Source & 0.273 & 0.196 & 0.188 & 0.219\\ \midrule
        AdaBN \cite{li2016revisiting} & 0.277 & 0.200 & 0.188 & 0.222\\
        OT \cite{wang2020train}& 0.199 & 0.166 & 0.153 & 0.173\\
        FS \cite{wang2020train} & 0.506 & 0.436 & 0.396 & 0.446\\
        \midrule
        SF-UDA$^{3D}$(SS) & 0.589 & 0.414 & 0.388 & 0.464\\ 
        SF-UDA$^{3D}$(MS-3) & \textbf{0.688} & \textbf{0.498} & \textbf{0.450} & \textbf{0.545}\\ 
        SF-UDA$^{3D}$(MS-5) & 0.657 & 0.479 & 0.427 & 0.521\\ 
        \midrule
        Target & 0.873 & 0.769 & 0.760 & 0.801\\ 
        \bottomrule
    \end{tabular}
    }
    \caption{Adaptation results: nuScenes$\rightarrow$KITTI}
    \label{tab:nusc2kitti}
    \vspace{-0.3cm}
\end{table}

\noindent\textbf{Implementation details.}
Our method is implemented in PyTorch, building upon the publicly-available PointRCNN 3D detector. More details can be found at the project website\footnote{https://github.com/saltoricristiano/SF-UDA-3DV}. For training and evaluating our method, we run all our experiments on a DGX-1 server (8 Nvidia Tesla V-100 GPUs) and on a Lambda Blade server (8 NVidia Quadro RTX 6000 GPUs). 
The source training is performed with the ADAM  optimization algorithm and one-cycle policy for 200 and 70 epochs for the RPN and RCNN respectively, with a batch size of 64 and 32 respectively, and with a maximum learning rate of $0.02$ as in \cite{shi2019pointrcnn}. Similarly, during the target fine-tuning we keep the same training setup with the only difference that we train the RPN for 100 epochs with a maximum learning rate of $0.002$.
During the scale search, we use a grid size parameter $\epsilon=0.3$ and we employ a stride $s=0.075$ along each axis. We thus obtain a solution space of dimension of $L=125$.
As for the tracking-by-detection module, we use the publicly-available code of \cite{weng2019baseline} and change the tracker hyper-parameters as follows: both the initialization and death thresholds are set to $2$ matched detections and missed detections respectively. Finally, regarding the pseudo-annotation procedure, we annotate four times the target dataset for each confidence threshold in the list $[0.05, 0.1, 0.2, 0.3]$ and use an NMS IoU threshold of $0.1$ to merge pseudo-annotations.
\vspace{-0.2cm}
\subsection{Comparison with the state-of-the-art}
\label{sec:exp-adaptation}

Our work is the first to tackle SF-UDA for 3D detection.
For the sake of evaluation against state-of-the-art methods, we still compare with the following methods:

\setlist{nolistsep}
\begin{itemize}[noitemsep]
    \item \noindent\textit{AdaBN}~\cite{li2016revisiting}: This is a state-of-the-art UDA method originally proposed for the classification tasks. We chose it because it is one of the few to operate in the source-free setting, differently from most previous approaches in UDA \cite{csurka2017domain,roy2019unsupervised,cariucci2017autodial}. Here, we adapted AdaBN to the detection task by updating the batch normalization (BN) source-model mean and variance statistics of the PointRCNN features to the target validation set.
    \item \noindent\textit{Few-Shot (FS) fine-tuning } \cite{wang2020train}: This method proposes to adapt the detector to the target domain by fine-tuning on a set of 10 randomly sampled annotated target data. Following \cite{wang2020train}, we ran it 5 times and reported the average performance.
    \item \noindent\textit{Output Transformation (OT)} \cite{wang2020train}: This is a weakly-supervised DA technique. It uses average sizes of objects in both the source and target domain to transform the predictions of the source model at test time over to the target data.
\end{itemize}

\begin{figure*}
    \centering
    \includegraphics[scale=0.4]{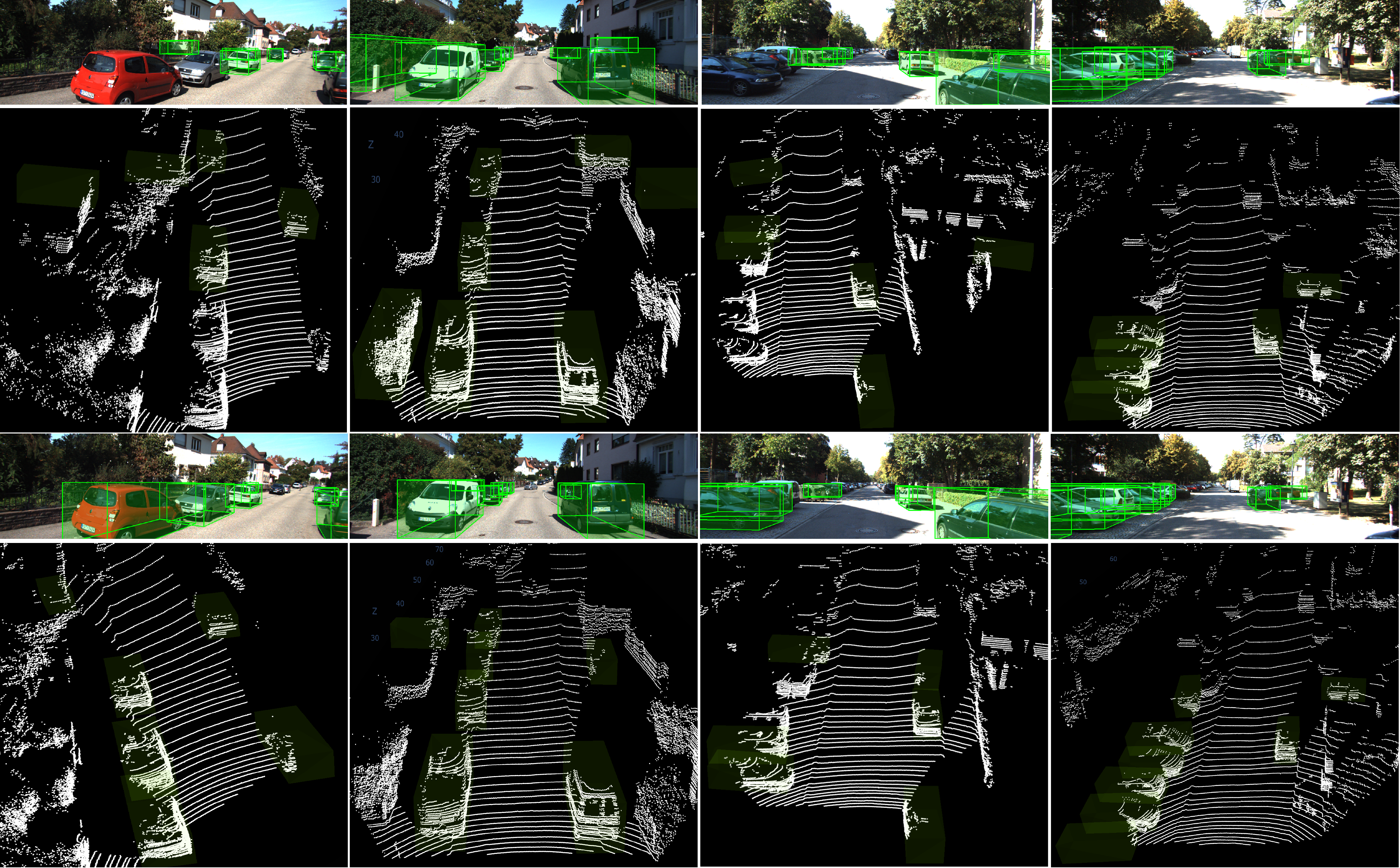}
    \caption{\textbf{Before (top) and after (bottom) adaptation on nuScenes$\rightarrow$KITTI.} After adaptation with MS-3, performance improves and more objects are detected.}
    \label{fig:detections}
    \vspace*{-0.3cm}
\end{figure*}

We compare SF-UDA$^{3D}$ in both the single-scale (SS) and top-$K$ multi-scale (MS-$K$) approaches, cf.\ Sec.~\ref{sec:annotation}. We consider two settings: (i) nuScenes as source and KITTI as target (nuScenes$\rightarrow$KITTI) and (ii) 
KITTI as source and nuScenes as target (KITTI$\rightarrow$nuScenes).
In order to be comparable with the literature on 3D object detection, we performed the experiments in the nuScenes$\rightarrow$KITTI setting by using the same split as in \cite{shi2019pointrcnn} and in the KITTI$\rightarrow$nuScenes setting by using the official nuScenes splits~\cite{nuscenes2019}. Note that in \cite{wang2020train} results are evaluated on different data splits which are not the official ones.

\begin{table}[t]
    \centering
        \resizebox{\columnwidth}{!}{
    \begin{tabular}{c|ccccc}
        \toprule
        \textbf{Method} & \textbf{AP-0.5} & \textbf{AP-1.0} & \textbf{AP-2.0} & \textbf{AP-4.0} & \textbf{Avg-AP}\\
        \midrule
        Source &  0.143 & 0.208 & 0.224 & 0.234 & 0.202\\ \midrule
        AdaBN \cite{li2016revisiting} & 0.144 & 0.208 & 0.224 & 0.234 & 0.203\\
        OT \cite{wang2020train} & 0.124 & 0.202 & 0.224 & 0.233 & 0.196\\
        FS \cite{wang2020train} & 0.170 & 0.211 & 0.235 & 0.250 & 0.216\\ 
        \midrule
        SF-UDA$^{3D}$(SS) & 0.136 & 0.260 & \textbf{0.290} & \textbf{0.308} & 0.249\\
        SF-UDA$^{3D}$(MS-3)\; & 0.203 & \textbf{0.266} & \textbf{0.290} & \textbf{0.308} & 0.267\\
        SF-UDA$^{3D}$(MS-5)\; & \textbf{0.211} & 0.264 & 0.288 & 0.307 & \textbf{0.268}\\ \midrule
        Target & 0.370 & 0.422  & 0.440 & 0.455 & 0.422\\
        \bottomrule
    \end{tabular}
    }
    \caption{Adaptation results: KITTI$\rightarrow$nuScenes}
    \label{tab:kitti2nusc}
    \vspace{-0.3cm}
\end{table}

The results on the nuScenes$\rightarrow$KITTI task are reported in Table~\ref{tab:nusc2kitti}; those on the KITTI$\rightarrow$nuScenes are shown in Table~\ref{tab:kitti2nusc}. In both cases, SF-UDA$^{3D}$ outperforms both OS~\cite{wang2020train} and FS~\cite{wang2020train}, although both of them use information from the target domain annotations.
Analysing the performance of different variations of our method we observe that in the nuScenes$\rightarrow$KITTI task (Table~\ref{tab:nusc2kitti}), the MS-3 version of SF-UDA$^{3D}$ is the best-performing method, gaining a $+0.08$ Avg-AP over SF-UDA$^{3D}$ (SS). 
Similarly, in the KITTI$\rightarrow$nuScenes task (Table~\ref{tab:kitti2nusc}) 
SF-UDA$^{3D}$ (MS-5) leads to an improvement of $+0.019$ Avg-AP compared to SF-UDA$^{3D}$ (SS). 
These results confirm the effectiveness of our scale-based pseudo-annotation approach and the importance of the combination of the top-$K$ solutions in the annotation procedure. 
Notably, AdaBN~\cite{li2016revisiting} is not effective on either the adaptation tasks. This may indicate that geometry-based methods are better-suited than feature-based methods for 3D LiDAR-based detection adaptation and that features may need more sophisticated DA approaches.
Additionally, to provide further insights on the results of Tables \ref{tab:nusc2kitti} and \ref{tab:kitti2nusc}, we also report and discuss the scaling parameters automatically selected by our method.
Considering experiments on the nuScenes$\rightarrow$KITTI task, SF-UDA$^{3D}$(SS) estimates the parameters $[1.30, 1.30, 1.15]$ along the axis X,Y and Z, respectively showing that KITTI data have to be upscaled along each axis to better match with nuScenes data. Similarly, SF-UDA$^{3D}$(MS-3) provides an indication in the upscaling direction and selects the scale parameters within the intervals $[1.15,1.30]$ along X, $[1.15,1.30]$ along Y and $[1.15,1.30]$ along Z.
Conversely, for the more challenging KITTI$\rightarrow$nuScenes setting, SF-UDA$^{3D}$(SS) computes the scaling parameters $[0.85, 0.70, 1.00]$, indicating that nuScenes objects point clouds should be downscaled along X and Y while keeping the original dimension along Z. Also the scale parameters adopted by the best performer SF-UDA$^{3D}$(MS-5) indicate downsampling for adaptation, by the ranges $[0.85,1.15]$ along X, $[0.70,0.85]$ along Y and $[0.85,1.00]$ along Z. Here the model finds the wider range in X beneficial.
Note that, in all of the above cases, the selected scales along the three axes are different. So further to scaling, SF-UDA$^{3D}$ implicitly learns to change the aspect ratios for domain adaptation.

%% file: datasets.tex

\begin{table*}[t]
 
    \centering
    \begin{tabular}{c|ccccccc}
       \toprule
        \textbf{Dataset} & \textbf{Samples} & \textbf{Max depth} & \textbf{Sensor} & \textbf{Channels} & \textbf{Resolution}  & \textbf{Mean points} & \textbf{Classes}\\    \midrule
        KITTI \cite{geiger2012we} & 15k & 70 m & HDL64E & 64 & $0.08\degree \times 0.4\degree$ & 16384 & 8\\ \hline
        nuScenes \cite{nuscenes2019} &  34k & 100 m & HDL32E & 32 & $0.08\degree \times 1.33\degree$ & 3808 & 23\\ \bottomrule
    \end{tabular}
    \caption{Datasets overview. Each dataset is acquired using sensors with different resolutions and numbers of channels. While nuScenes uses the original maximum depth, KITTI provides pre-filtered LiDAR data with a maximum depth of 70 m.}
    \label{tab:sensors}
    \vspace{-0.2cm}
\end{table*}

\begin{figure}[t]
\centering
\begin{subfigure}{0.45\columnwidth}
  \centering
  \includegraphics[width=0.85\textwidth]{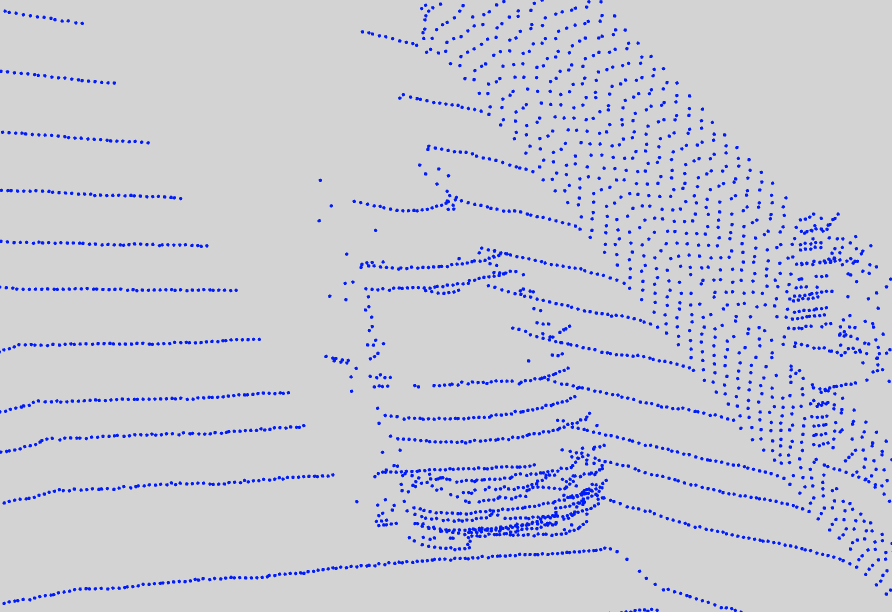}
  \caption{KITTI \cite{geiger2013vision}}
  \label{fig:kitti}
\end{subfigure}
\begin{subfigure}{0.45\columnwidth}
  \centering
  \includegraphics[width=0.9\textwidth]{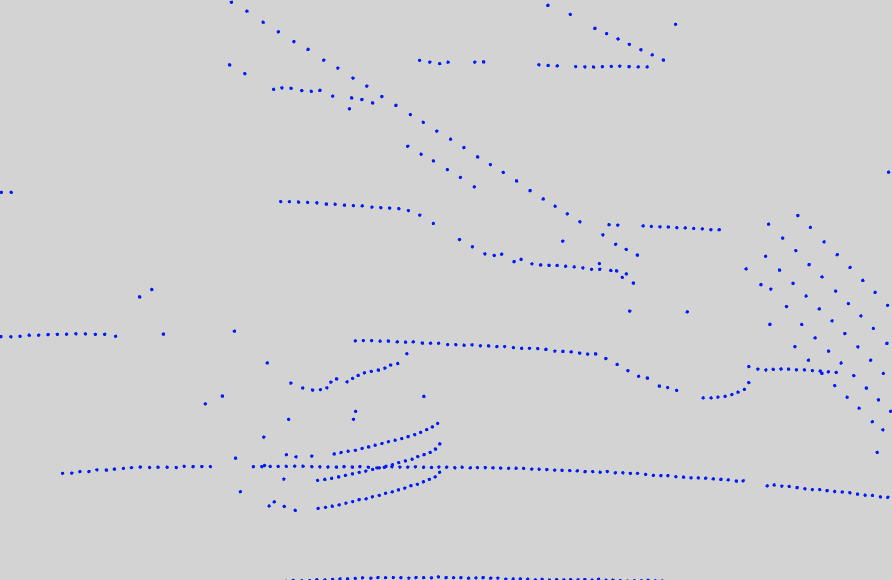}
  \caption{nuScenes \cite{nuscenes2019}}
  \label{fig:nuscenes}
\end{subfigure}%
\caption{Example of cars from the KITTI and nuScenes datasets.}
\label{fig:comparison}
\vspace{-0.5cm}
\end{figure}

\noindent\textbf{Datasets.}
The KITTI object detection benchmark dataset \cite{geiger2012we, geiger2013vision} has been acquired in Karlsruhe, Germany and is composed of 7481 training images, divided into 3712 training samples, 3769 validation samples and 7518 test images.\\
The nuScenes dataset \cite{nuscenes2019}, acquired in Boston (USA) and Singapore, is $\sim2.3$ times larger, composed of 1000 driving sequences, for a total of 34149 images, divided into 28130 training samples and 6019 validation samples (which we treat as test samples), and it has both LiDAR scans and RGB images.
The datasets differ under three main aspects:
\begin{itemize}
    \item \textit{Sensors.} Different sensors were used for data acquisition, as summarized in Table \ref{tab:sensors}. These sensors sample points differently in terms of density (\textit{i.e.} number of points), temporal frequency, spatial resolutions and ranges. Fig.~\ref{fig:comparison} illustrate example differences, which affect the performances of the 3D object detectors.
   \item \textit{Environmental conditions.} Being acquired in diverse countries, the datasets depict objects of different shapes and sizes.
   Cars, which we target in this work, change much in these two aspects due to the differences between Germany and USA.
   %
    \item \noindent\textit{Dataset pre-processing.}
    The authors of the datasets made different choices for the data collection, annotation and filtering. For instance, in KITTI, only objects within $70m$ are annotated, while in nuScenes objects up to $100m$ have ground truth annotations. Another difference is the range where objects are annotated. Indeed, in KITTI, only objects visible in the front camera view are annotated, while in nuScenes also objects are annotated in the entire $360\degree$ surrounding space. In all our experiments, we consider the points which are visible from the frontal RGB-camera viewpoint (\eg the CAM-FRONT in nuScenes and the rectified camera space for KITTI).
    %
    
\end{itemize}


%% file: ablation.tex
\subsection{Ablation studies}
\label{sec:ablation}
\vspace{-0.2cm}
Here we present the results of the ablation on each components of our method.

\noindent \textbf{Scaling parameters.}
First we investigate the importance of the scaling parameters in the single-scale (SS) setup of SF-UDA$^{3D}$ during the pseudo-annotation phase. We consider the nuScenes$\rightarrow$KITTI task and compare our results with three baselines:
\begin{itemize}
\item \textit{No score}: we remove from SF-UDA$^{3D}$ the temporal-coherence tracking-based scores. This results in the random sampling of $\omega$ in the range $[0.7,1.3]$ along each axis during the pseudo-annotation phase and quantifies the mere scale-augmentation.
\item \textit{No scale}: we remove from SF-UDA$^{3D}$ the scale-transformations altogether. So the target dataset is simply pseudo-annotated by the source model \emph{as is}.
\item \textit{Supervised scale (Sup-scale)}: we make the single-scale SF-UDA$^{3D}$ weakly-supervised, by providing it with the ground-truth scale differences between the target and source average object sizes.
\end{itemize}

Table \ref{tab:abl-scaling} reports the results of our evaluation, comparing different scaling methods in term of Avg-AP and reporting the selected scales.
We observe from Table \ref{tab:abl-scaling} that \textit{No scale} is a poor adaptation strategy, probably because there is a significant average scale difference between KITTI and nuScenes. Among the three ablation variants, \textit{No score} is the worst, a bit surprisingly. While data augmentation is mostly a positive tool, here it shows that it is important to augment at the relevant scales, which we determine by tracking. \textit{Sup-scale} respects the intuition and is slightly above our single-scale (SS) SF-UDA$^{3D}$, since knowing the ground-truth is important. However the improvement is marginal, which confirms that the estimation scale-transformation parameters is effective.



\begin{table}[t]
    \centering
   \resizebox{\columnwidth}{!}{
    \begin{tabular}{c|c@{}c@{}c|c}
        \toprule
        \multirow{2}{*}{\textbf{Method}}
      & \multicolumn{3}{c|}{\textbf{Selected Scale}}  &       \multirow{2}{*}{\textbf{Avg-AP}}\\
      & \textbf{X} & \textbf{Y} & \textbf{Z} &       \\
         \midrule
        No scale & 1.00 & 1.00 & 1.00 & 0.306\\
        No score & \;$[0.70,1.30]$\; & \;$[0.70,1.30]$\; & \;$[0.70,1.30]$\; & 0.223\\
        SS & 1.30 & 1.30 & 1.15 & 0.464\\
        Sup-scale & 1.18 & 1.09 & 1.16 & \textbf{0.499}\\
        \bottomrule
    \end{tabular}
}
    \caption{Ablation results: different scaling parameters.}
    \label{tab:abl-scaling}
    \vspace{-0.5cm}
\end{table}

\noindent\textbf{Scale selection metric.}
In a second ablation experiment, we compare possible variants of unsupervised metrics to assess the quality of tracks (and thus to identify the most suitable scale transformations) on both the considered adaptation settings. In more details, we take inspiration from the work of \cite{xu2016libsvx} on benchmarking supervoxels and compare the applicable unsupervised metrics, namely the Time-Extension (TEX) and the Mean Volume Variation (MVV) -- note that \cite{xu2016libsvx} introduces MSV, applicable to images, which we generalize to the point cloud volumes with MVV. TEX measures the temporal extent of tracking across frames, since intuitively a more stable tracking algorithm generates longer tracks. For MVV we compare the bare metric, as well as our proposed extension MVV$^*$ with the penalty below a minimum-length track.
\begin{table}[ht]
    \centering
    \begin{tabular}{c|cc}
        \toprule
                \multirow{2}{*}{\textbf{Metric}}       &  \multicolumn{2}{c}{\textbf{Avg-AP}}\\
      & nuScenes$\rightarrow$KITTI & KITTI$\rightarrow$nuScenes     \\
          \midrule
        TEX \cite{xu2016libsvx} & 0.030 & 0.229 \\
        MVV \cite{xu2016libsvx} & \textbf{0.488} & 0.125\\
        MVV$^*$ & 0.464 & \textbf{0.249}\\
        \bottomrule
    \end{tabular}
    \caption{Ablation results: different scoring metrics.}
    \label{tab:scale_selection}
    \vspace{-0.3cm}
\end{table}

From Table~\ref{tab:scale_selection}, we see that TEX is not a suitable metric, though. In the realm of point clouds, longer tracks may correspond to wrong matches over time. In the Table, it shows that MVV is slightly better for the adaptation nuScenes$\rightarrow$KITTI, but MVV$^*$ greatly outperforms it for the adaptation KITTI$\rightarrow$nuScenes. So the penalty terms plays an important role in the scoring over the noisier and more difficult nuScenes dataset.

\noindent\textbf{Assessing pseudo-annotations by scaling just.}
In this final ablation study, we target to measure the quality of the pseudo-labels due to scaling transformations and to investigate why the combination of multiple scales is superior to the single-scale (e.g.\ MS-3 Vs.\ SS) pseudo-labelling.
To this goal, we focus on estimating the best scales by temporal coherency. Then we consider each of the three best (1st, 2nd and 3rd best) and re-scale the target point-cloud according to it. Finally, we detect with the original source model $\Phi_S$, without fine-tuning. Compared to the full pipeline of Fig.~\ref{fig:full}, this study excludes step 4 of the pipeline. 
The results are reported in Table~\ref{tab:top-k}. Re-scaling the point cloud according to each of the three best scales improves the performance of the source model considerably. In fact, the source model passes from a performance of $0.202$ Avg-AP on the non-rescaled point-cloud (\textit{No scale}) to $0.370$ Avg-AP in the case of the 1st best, which is 83\% better. Also, re-scaling according to the ground-truth annotation statistics (\textit{Sup-scale} entry in the Table) is understandably better. Finally, note that the three best scales resize the point-cloud in the same direction, by approximately similar upscaling factors, but with different aspect ratios. This may account for the better performance of MS-3 Vs.\ SS.

\begin{table}[t]
    \centering
    \begin{tabular}{c|ccc|c}
        \toprule
        \multirow{2}{*}{\textbf{Scale}}
      & \multicolumn{3}{c|}{\textbf{Selected Scale}}  &       \multirow{2}{*}{\textbf{Avg-AP}}\\
      & \textbf{X} & \textbf{Y} & \textbf{Z} &       \\
         \midrule
        No scale & 1.00 & 1.00 & 1.00 & 0.202 \\
        $1^{st}$ best & 1.30 & 1.30 & 1.15 & 0.370 \\
        $2^{nd}$ best & 1.15 & 1.30 & 1.30 & 0.394\\
        $3^{rd}$ best & 1.30 & 1.15 & 1.15 & 0.369\\
        Sup-scale & 1.18 & 1.09 & 1.16 & \textbf{0.435} \\
        \bottomrule
    \end{tabular}

    \caption{nuScenes$\rightarrow$KITTI setting: quality of pseudo-annotations by using the best 3 scored scaling parameters. The results differ from Table \ref{tab:abl-scaling} since here we measure the quality of pseudo-annotations before the fine-tuning step.}
    \label{tab:top-k}
    \vspace{-0.5cm}
\end{table}

%% file: conclusions.tex
\vspace{-0.25cm}
\section{Conclusions}
\vspace{-0.2cm}
We have proposed SF-UDA$^{3D}$, the first  Source-Free Unsupervised Domain Adaptation approach for 3D LiDAR-based detection. Leveraging on pseudo-annotations, motion coherence and reversible scale-transformations, our method is capable to adapt a LiDAR-based detector trained on source data to a new target domain where only unlabelled pointclouds are available. Domain adaptation experiments conducted on the nuScenes$\rightarrow$KITTI and KITTI$\rightarrow$nuScenes tasks have shown that SF-UDA$^{3D}$ outperforms state of the art weakly supervised and few-shot supervised methods by a large margin. In future work, we plan to extend the method beyond cars, to the other object classes, to investigate the use of depth maps for adaptation \cite{pilzer2019refine, PilzerPAMI, ricci2018monocular} and to continue the investigation of unsupervised metrics for the scoring procedure.